# Probabilities-Informed Machine Learning

Mohsen Rashki
Department of Architectural Engineering, University of Sistan and Baluchestan, Zahedan, Iran
Email address: Mrashki@eng.usb.ac.ir

**Abstract**
Machine learning (ML) has emerged as a powerful tool for tackling complex regression and classification tasks, yet its success often hinges on the quality of training data. This study introduces an ML paradigm inspired by domain knowledge of the structure of output function, akin to physics-informed ML, but rooted in probabilistic principles rather than physical laws. The proposed approach integrates the probabilistic structure of the target variable—such as its cumulative distribution function—into the training process. This probabilistic information is obtained from historical data or estimated using structural reliability methods during experimental design. By embedding domain-specific probabilistic insights into the learning process, the technique enhances model accuracy and mitigates risks of overfitting and underfitting. Applications in regression, image denoising, and classification demonstrate the approach's effectiveness in addressing real-world problems.

## 1. Introduction

Over the last decade, Machine Learning (ML) has emerged as a transformative tool for solving complex prediction and classification problems [1], [2]. As a natural evolution of traditional regression methods [3], ML models such as Support Vector Regression (SVR) [4] and Artificial Neural Networks (ANN) [5] have been developed to handle non-linear relationships and high-dimensional datasets [6] with increasing accuracy and robustness.

For instance, SVR has proven to be a robust regression tool because it can generalize well with limited data and capture nonlinear relationships using kernel functions [7]. Similarly, ANN, inspired by the neural architecture of the human brain, has become foundational to ML [5]. Typically, these methods use inputs (**X**) and outputs ($Y$) to construct surrogate models that aim to minimize the difference between the predicted and actual output values. These models have found applications across diverse fields, including engineering, medicine, and economics, demonstrating their versatility and potential [8], [9], [10].

In many real-world applications, additional prior information regarding the output model can be leveraged to enhance its accuracy and robustness [11] [12]. For instance, in physical systems, knowledge of the governing laws of physics has been successfully incorporated into ML by developing physics-informed neural networks (PINNs) [13], leading to improved efficiency and accuracy in prediction tasks [14].

In addition to physical laws, probabilistic information about the structure of the problem may also exist in practical scenarios [15]. Moreover, in many systems, the output variable is inherently probabilistic, necessitating models to approximate the probabilistic structure of the output [16]. Methods such as the Gaussian Processes (GPs) [17] [18], Bayesian neural networks [19], quantiles [20] and generative modeling [21] have shown great potential for integrating such probabilistic issues in ML. Furthermore, as demonstrated in [15], [22] and [23], probabilities-informed modeling enhances robust decision-making and ensures reliable predictions, particularly in complex systems (see also [24] and [25] where the concept of failure probability is utilized to improve the training process in PINNs). These findings underscore the critical role of leveraging the complete probabilistic structure of the output function during model selection and training in ML.

This paper develops probabilities-informed modeling as an explicit framework for incorporating probabilistic information into ML to enhance its effectiveness. In the next section, we discuss how probabilistic information can be effectively utilized to select an appropriate model for prediction tasks and propose strategies for deriving the probabilistic structure of the output function *Y*. Section 3 introduces a framework for training ML models by integrating probabilistic information. In Section 4, we validate the proposed method through various applications, and finally, Section 5 presents the conclusions and potential future directions.

## 2. Problem Statement and Study Motivation

In ML, overfitting and underfitting are fundamental challenges that impair model generalization and prediction reliability. Overfitting occurs when a model captures noise or overly specific patterns in the training data, reducing its ability to generalize to unseen data. Conversely, underfitting arises when the model fails to capture the inherent complexity of the data, leading to suboptimal predictive performance.

These issues are particularly acute in cases where the target variable *Y* exhibits complex or probabilistic behavior, such as multi-modal distributions or inherent uncertainties. Relying solely on error-based metrics like Root Mean Squared Error (RMSE) for model selection often leads to the choice of an inadequate model, either due to overfitting or underfitting, failing to represent the underlying data distribution accurately.

For instance, consider a dataset with $n$=10 input features $\mathbf{X} = [X_1, X_2, \dots, X_{10}]$ and corresponding outputs $\boldsymbol{Y}$. The scatter plot of data is presented in Fig. 1.

The objective is to train a predictive model $\hat{Y} = f(\boldsymbol{X})$ to approximate the relationship between $X$ and $Y$**,** explore polynomial models of order $i$ (e.g., $i$=1,2,…,5).

When the dataset $(\boldsymbol{X}, \boldsymbol{Y})$ is the sole source of information and the problem is treated as a black-box problem (e.g., in the absence of domain information about the output *Y*), RMSE is typically employed to evaluate model performance and select the optimal polynomial order.

In this scenario, selecting a polynomial of order $i$=2 (Model #2) might appear reasonable due to its balance between simplicity, and error minimization, leaving room for ambiguity regarding how well the model captures the underlying structures of $Y$.

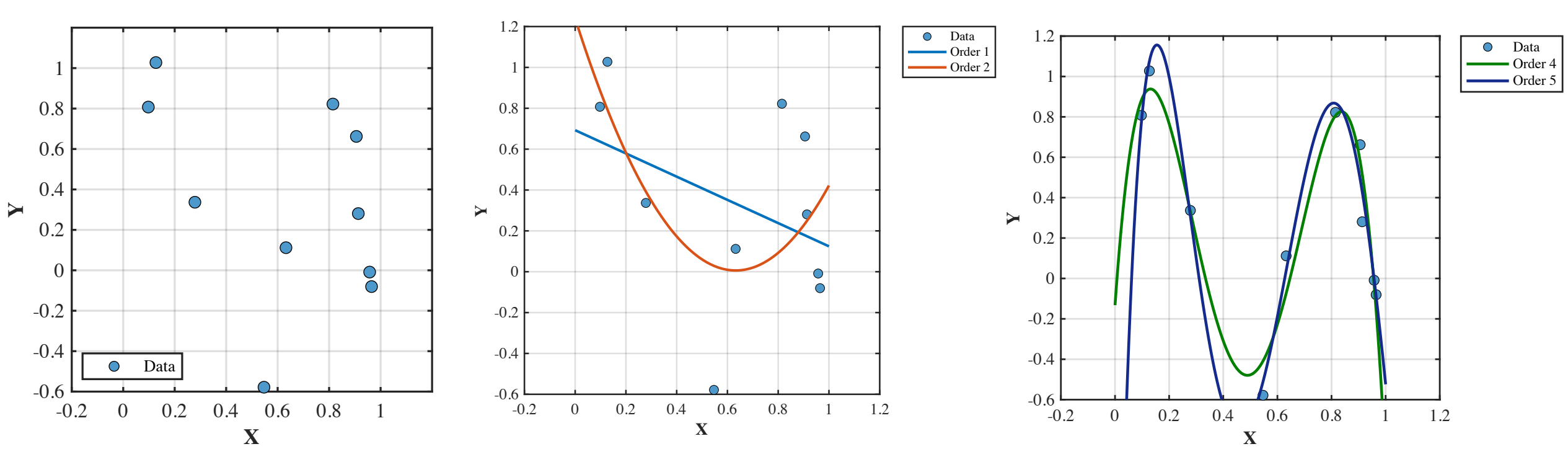

Fig. 1. Illustration of available data alongside fitted polynomial models of varying complexity

### 2.1. The Value of Probabilities-Informed Modeling

Recent research has demonstrated that incorporating probability-informed modeling improves mobile network traffic analysis [15] and time-dependent reliability results [23] [22] in intricate systems.

In the context of the previously discussed problem, we explored the limitations of selecting a model based solely on the dataset $(\boldsymbol{X}, \boldsymbol{Y})$. In this subsection, suppose a scenario in which some additional probabilistic information about X and $Y$ is available.

Let $\mathbf{X} \in \mathbb{R}^n$ follow a known distribution $P_X(\boldsymbol{x})$ and $Y$ be governed by a conditional probability distribution $P_Y(y|\mathbf{X} = \boldsymbol{x})$. Assume the marginal distributions $P_X$ and $P_Y$, as well as the empirical cumulative distribution function (CDF) of $\boldsymbol{Y}$, are available as shown in Fig. 2.

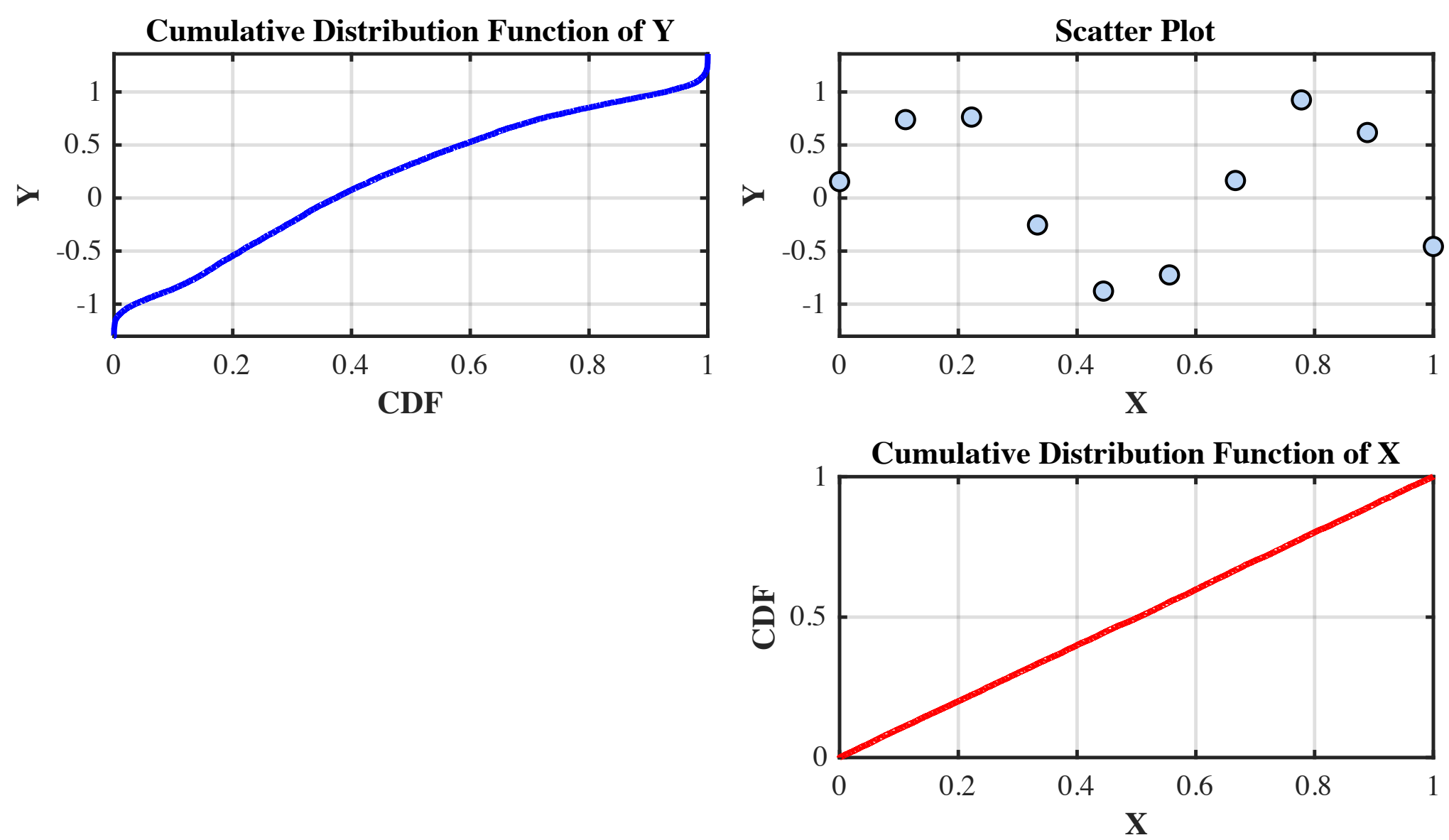

Fig. 2. The dataset $(\boldsymbol{X}, \boldsymbol{Y})$ and full probabilistic structure of $Y$

This additional information enables us to evaluate model suitability beyond RMSE. Consider two polynomial models: $f_2(\boldsymbol{x})$ (order 2) and $f_5(\boldsymbol{x})$ (order 5), as shown in Fig. 3. Using the known distribution $P_X(\boldsymbol{x})$, we can generate a large number of synthetic samples of X (uniform samples in the mentioned case), compute the corresponding $\hat{Y}$ values for each model, and construct histograms or empirical CDFs of the predicted $Y$. Let $\hat{F}_{f_2}(y)$ and $\hat{F}_{f_5}(y)$ denote the empirical CDFs of predictions from the two models, and let $F_Y(y)$ be the empirical CDF of the $Y$ (i.e., available probabilistic information). By comparing $\hat{F}_{f_2}(y)$ and $\hat{F}_{f_5}(y)$ against $F_Y(y)$, we can assess how well each model captures the probabilistic structure of $Y$. These explanations are illustrated in Fig 3 in detail.

Fig. 3-D illustrates that the predictions from $f_2(\boldsymbol{x})$ fail to match $F_Y(y)$, despite achieving a reasonable RMSE in the training process, indicating that Model #2 underestimates the true complexity of $Y$. In contrast, predictions from $f_5(\boldsymbol{x})$ align closely with $F_Y(y)$, suggesting that polynomial order 5 better captures the probabilistic structure of $Y$ while maintaining acceptable RMSE.

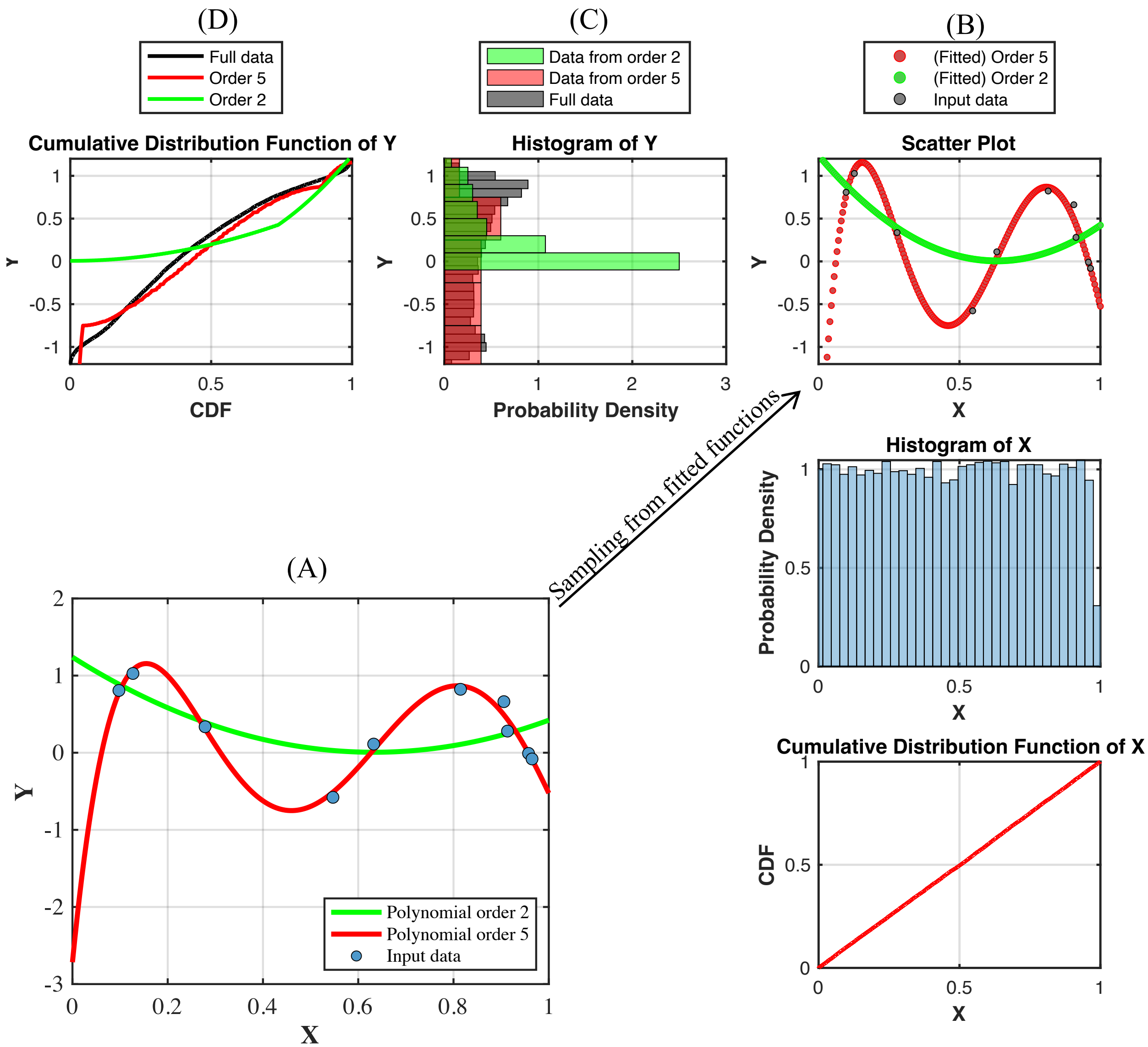

Fig. 3. (A) Two candidate prediction models, $\hat{F}_{f_2}(y)$ and $\hat{F}_{f_5}(y)$; (B) Sample generation based on the distribution of X and evaluation of $\hat{Y}$, (C) Derivation of the PDF of $\hat{Y}$, and, D) Calculation of the CDF of $\hat{Y}$

The proposed explanation underscores that incorporating probability information about the structure of $Y$ provides a robust framework for model selection. By incorporating the known marginal distribution $P_X(\boldsymbol{x})$ and the conditional distribution $P_Y(y|\mathbf{X}=\boldsymbol{x})$, we transform the problem from a purely data-driven black box into a semi-informed problem. By combining this probabilistic information with conventional metrics like RMSE, as shown in Fig. 4, we can identify models that not only minimize error but also faithfully represent the underlying probabilistic data structures (e.g., Polynomial order 5, in this example).

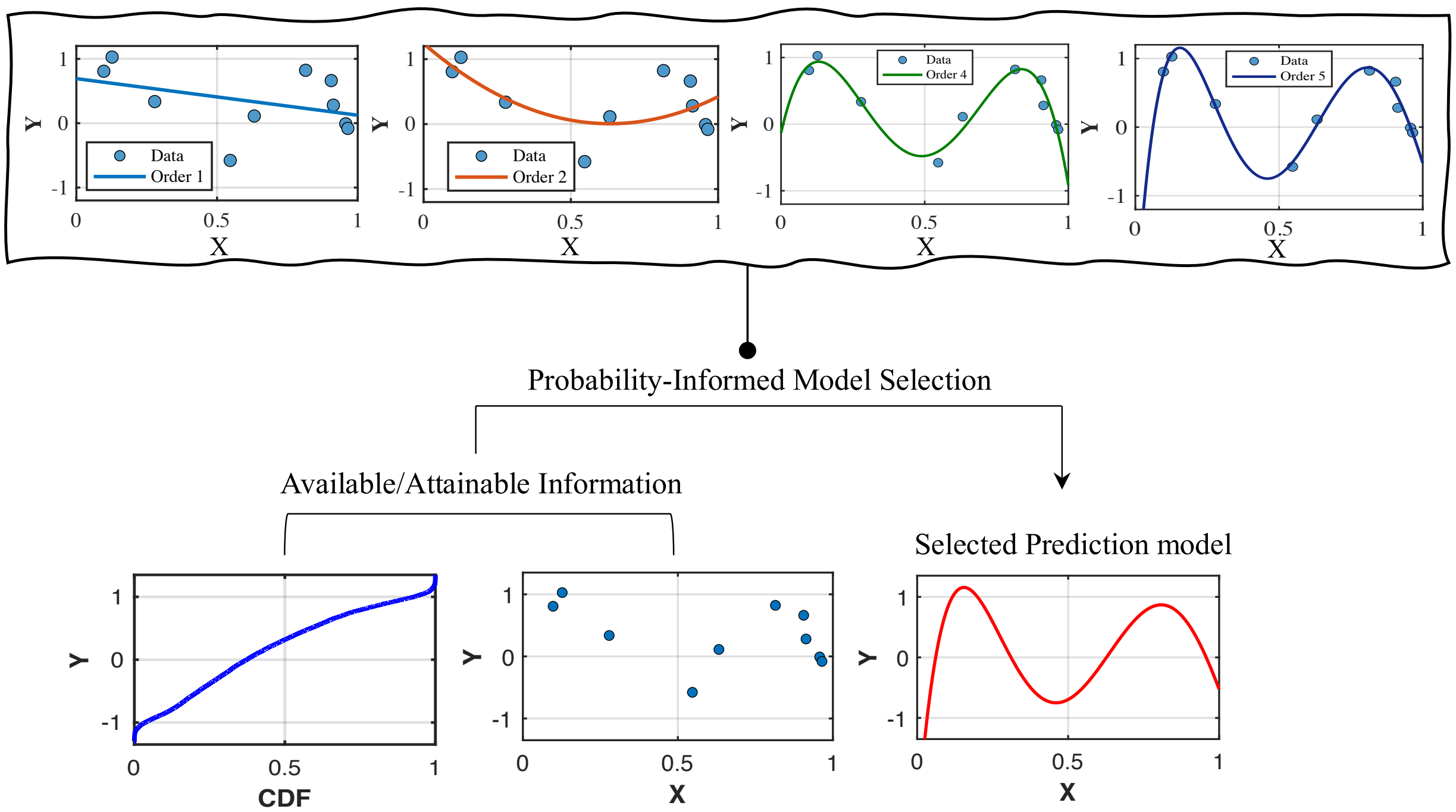

Fig. 4. Model selection based on the dataset $(\boldsymbol{X}, \boldsymbol{Y})$ and the probabilistic behavior of $\boldsymbol{Y}$

### 2.2. Deriving the Probabilistic Structure of Output $\boldsymbol{Y}$

In this section, we focus on deriving the CDF $F_Y(y)$ as the probabilistic structure of $Y$. The derived CDF allows us to identify regions where $Y$ is more likely or less likely to occur, and therefore serves as a benchmark for evaluating candidate models $\hat{Y} = f(\mathbf{X})$.
By comparing the empirical CDF of model predictions $\hat{F}_Y(y)$ with $F_Y(y)$, we can determine how well the model captures the true probabilistic behavior of $Y$.

#### 2.2.1. Empirical Derivation of the CDF of $\boldsymbol{Y}$ (In the Presence of Data $\boldsymbol{Y}$)

In many real-world scenarios, empirical data enable the estimation of the $P_X(\boldsymbol{x})$ and CDF $F_Y(y)$, even when the relationship between $Y$ and its influencing factors $\mathbf{X}$ is unknown or sparsely documented. For instance:

- **Engineering**: The number of cycles to failure for materials under repeated loading is empirically characterized, allowing CDF estimation, though its dependence on material and loading parameters remains unclear.
- **Economics**: Selling time or price of properties is recorded, enabling CDF estimation despite limited knowledge of influencing market and property features.
- **Manufacturing**: Time-to-failure for components is measured empirically, facilitating CDF estimation, while its relationship to production and operational factors remains uncertain.
- **Energy Systems**: Energy consumption patterns are observed and modeled empirically, though the connection to temperature, seasonality, and human behavior often lacks data and explicit definition.
- **Medical Research**: Survival times of patients are empirically modeled, though their relationship with treatment and patient-specific factors is complex and data-limited.

These examples highlight scenarios where $F_Y(y)$ can be estimated directly, while the functional dependency on **X** requires innovative modeling approaches.

Empirical data in these cases are often derived from observed measurements or experiments, where the output variable $Y$ is recorded. For a set of thresholds $\{y_k\}_{k=1}^{K}$ (where $y_k$ represents a specific value within the range of $Y$), $F_Y(y_k) = \mathbb{P}(Y \leq y_k)$ can be estimated directly from the data using non-parametric methods, such as the empirical CDF:

$$F_Y(y_k) = \frac{1}{n}\sum_{i=1}^{n} I_{(y_i \leq y_k)}, \qquad (1)$$

where $y_1$, $y_2$,…, $y_n$ are the observed values of $Y$, and $I_{(y_i \leq y)}$ is the indicator function, equal to 1 if $y_i \leq y_k$ and 0 otherwise. Then, the final CDF $F_Y(Y)$ for any $y$ can be expressed as:

$$F_Y(y) = \begin{cases} 0 & \text{if } y < y_{min}, \\ F_Y(y_k) & \text{for } y_k \leq y < y_{k+1} (\text{interpolated}), \\ 1 & \text{if } y \geq y_{max}, \end{cases} \qquad (2)$$

where $y_{min}$ and $y_{max}$ are the minimum and maximum thresholds.

#### 2.2.2. CDF Estimation of *Y* (In the Absence of Data *Y*) Through Structural Reliability Theory

In cases where empirical data for $Y$ is unavailable, the CDF can still be estimated by leveraging a properly designed set of experiments (DOE) and employing structural reliability methods. These approaches provide probabilistic insights into $Y$ based on the known or assumed distributions of input variables **X**. In this context, the key challenge lies in deriving reliable probabilistic results while ensuring the sampled data are designed to adequately represent the variability and dependency structure of **X**, as required by the DOE.

Let $\mathbf{X} = [X_1, X_2, \dots, X_n]$ represent the vector of input variables. Suppose **X** is distributed according to a joint probability density function $P_X(\boldsymbol{x})$. Then, the CDF of $Y$, which gives the probability that $Y$ takes a value less than or equal to $y$, is given by:

$$F_Y(y) = \mathbb{P}(Y \leq y) = \int_{-\infty}^{y} P_Y(y)dy = \int_{Y \leq y} P_X(\boldsymbol{x})d\boldsymbol{x}. \qquad (3)$$

Equation (3) represents a well-known probability integral in structural reliability theory and can be used to derive probabilities for required thresholds $\{y_k\}_{k=1}^{K}$.

Recognizing that estimating extremely small probabilities can be disregarded in this context (as accurately predicting extreme values may not be necessary), various numerical and approximate methods from structural reliability theory can be utilized to estimate $F_Y(y)$ (e.g., Monte Carlo simulation [26], Importance sampling [27], Subset simulation [28], Soft Monte Carlo simulation [29], etc.). Notably, the same input data **X** can serve a dual purpose: it supports reliability analysis and simultaneously provides a foundation for training models based on the dataset ($\boldsymbol{X}, \boldsymbol{Y}$) (See Fig. 5). As a straightforward approach, random sampling based on the Monte Carlo method (MCS) can be used to estimate the CDF and desired dataset for proper machine training.

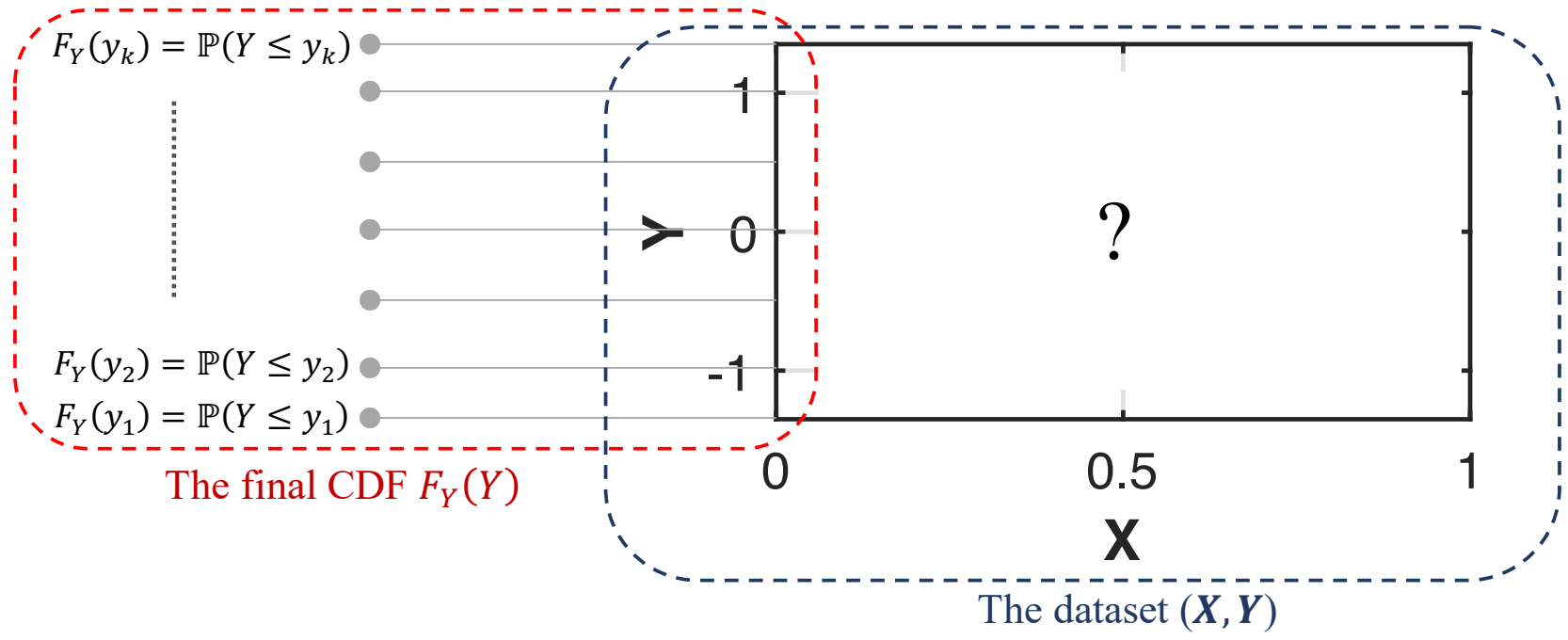

Fig. 5. In the absence of empirical data $Y$, a dual-purpose DOE is required to construct an appropriate dataset $(\boldsymbol{X}, \boldsymbol{Y})$ and to derive the CDF of $Y$ using a reliability method (estimating very small probabilities is unnecessary when precise prediction of extreme values is not essential)

## 3. Probabilities-Informed Machine Learning (PRIML)

The core idea of Physics-Informed Machine Learning (PIML) is to incorporate physical laws or constraints directly into ML models to enhance their predictive accuracy and consistency with real-world phenomena. Noting the point that the value of probabilities-informed modeling has formerly been discussed for specific applications (see [15], [23] and [22]), this study presents Probabilities-Informed Machine Learning (PRIML) as a generalized framework that integrates probabilistic knowledge into ML, focusing on both accuracy and capturing the true distribution of the data. As a result, unlike traditional models that prioritize minimizing prediction errors (e.g., RMSE), PRIML introduces a probabilistic constraint to ensure the predicted outputs $\hat{Y}$ align with the distributional properties of the true outputs $Y$. This is achieved through two complementary components:

- RMSE: To minimize individual prediction errors and achieve accuracy at the point level.
- CDF-based distance measure: To align predicted and true distributions of output $Y$, ensuring consistency with the underlying probabilistic structure.

Using the proposed insight, developments in reliability theory can be employed to train better models, whether empirical (probabilistic) data is available or when dealing with complex black-box problems. By leveraging reliability-based approaches, we can obtain more reliable results that provide critical insights for improving model accuracy, stability, and overall performance, even in the absence of physical constraints or detailed models.

### 3.1. Mathematical Formulation

Let $\boldsymbol{X} \in \mathbb{R}^n$ represent the input features and $Y \in \mathbb{R}$ represent the output variable, which follows a known probability distribution $P_Y(y)$. The goal is to train a model $f_\theta(X)$ with parameters $\boldsymbol{\theta}$, such that:

1. $f_\theta(X)$ accurately predicts $Y$, minimizing individual prediction errors (low RMSE)
2. The distribution of $f_\theta(X)$, denoted $\hat{P}_Y(y)$, matches the true distribution $P_Y(y)$.

To achieve this, we define a loss function that balances point-wise accuracy and distributional alignment:

$$\mathcal{L}(\theta) = \alpha \cdot \mathcal{L}_{\text{Data}}(\theta) + \beta \cdot \mathcal{L}_{\text{Prob}}(\theta), \qquad (4)$$

where $\mathcal{L}_{\text{Data}}(\theta)$ represents the data-driven loss term that focuses on minimizing prediction errors (e.g., mean squared error denoted as MSE) and $\mathcal{L}_{\text{Prob}}(\theta) = \mathcal{L}_{\text{Prob}}\left(P_Y(y), \hat{P}_Y(y),\right)$ quantifies the alignment between the predicted and true cumulative distributions. The coefficients $\alpha$ and $\beta$ balance these terms, $\alpha$ emphasizing point-wise prediction accuracy and $\beta$ prioritizing distributional alignment to capture the system's probabilistic behavior.

The data-driven loss term ensures point-wise prediction accuracy:

$$\mathcal{L}_{\text{Data}}(\theta) = \frac{1}{N}\sum_{i=1}^{N}(Y_i - f_\theta(X_i))^2, \qquad (5)$$

where $N$ is the number of training samples.

To measure the alignment between the true distribution $P_Y(y)$ and the predicted distribution $\hat{P}_Y(y)$, we define the cumulative distribution-based distance as loss function $\mathcal{L}_{\text{Prob}}(\theta)$. Let the empirical CDFs of $Y$ and $\hat{Y}$ (the predictions) be $F_Y(y) = \mathbb{P}(Y \leq y)$, and $\hat{F}_Y(y;\theta) = \mathbb{P}(f_\theta(X) \leq y)$. Then the loss $\mathcal{L}_{\text{Prob}}(\theta)$ is expressed as:

$$\mathcal{L}_{\text{Prob}}(\theta) = \int_{-\infty}^{\infty}\left|F_Y(y) - \hat{F}_Y(y;\theta)\right|dy. \qquad (6)$$

In practice, the integral is approximated using a finite set of thresholds $\{y_1, y_2, \ldots, y_k\}$:

$$\mathcal{L}_{\text{Prob}}(\theta) \approx \sum_{j=1}^{k}\left|F_Y(y) - \hat{F}_Y(y;\theta)\right|. \qquad (7)$$

Therefore, the proposed loss function can be replaced with alternative CDF distance measures ($D(F_Y, \hat{F}_Y)$)

$$\mathcal{L}_{\text{Prob}}(\theta) = D(F_Y, \hat{F}_Y), \qquad (8)$$

where the Bhattacharyya Distance:

$$D_B(F_Y, \hat{F}_Y) = -\log\int\sqrt{F_Y(y)\hat{F}_Y(y)}\,dy \qquad (9)$$

or Kullback-Leibler (KL) divergence

$$D_{KL}(F_Y, \hat{F}_Y) = \int F_Y(y)\log\left(\frac{F_Y(y)}{\hat{F}_Y(y)}\right)dy, \qquad (10)$$

or, based on the in-hand problem and employed machine learning approach, other distance measures (e.g., $D_{JS}(F_Y, \hat{F}_Y)$ as Jensen-Shannon Divergence) [30] [31] [32] can be used to present the total loss function. A comprehensive overview of potential loss functions and evaluation measures, which may apply to the proposed approach, can be found in [16].

In the general case where other prior knowledge is involved, the total loss function integrates terms for proposed objectives (e.g., prediction accuracy and distributional consistency) and other constraints. For instance, the loss function with physics-based constraints can be expressed as:

$$\mathcal{L}(\theta) = \alpha\cdot\mathcal{L}_{\text{Data}}(\theta) + \beta\cdot\mathcal{L}_{\text{Prob}}(\theta) + \gamma\cdot\mathcal{L}_{\text{Physics}}(\theta), \qquad (11)$$

where $\gamma$ is a weighting factor to balance the contributions of the last term and $\mathcal{L}_{\text{Physics}}(\theta)$ is the physics-based loss term that penalizes violations of physical constraints or governing equations

$$\mathcal{L}_{\text{Physics}}(\theta) = \frac{1}{N}\sum_{i=1}^{N}\left\|\mathcal{R}(\hat{Y}, \mathbf{X}, \theta)\right\|^2, \qquad (12)$$

in which $\mathcal{R}$ is a residual function encoding the physical laws. Fig. 6 describes the flow of information in the proposed approach.

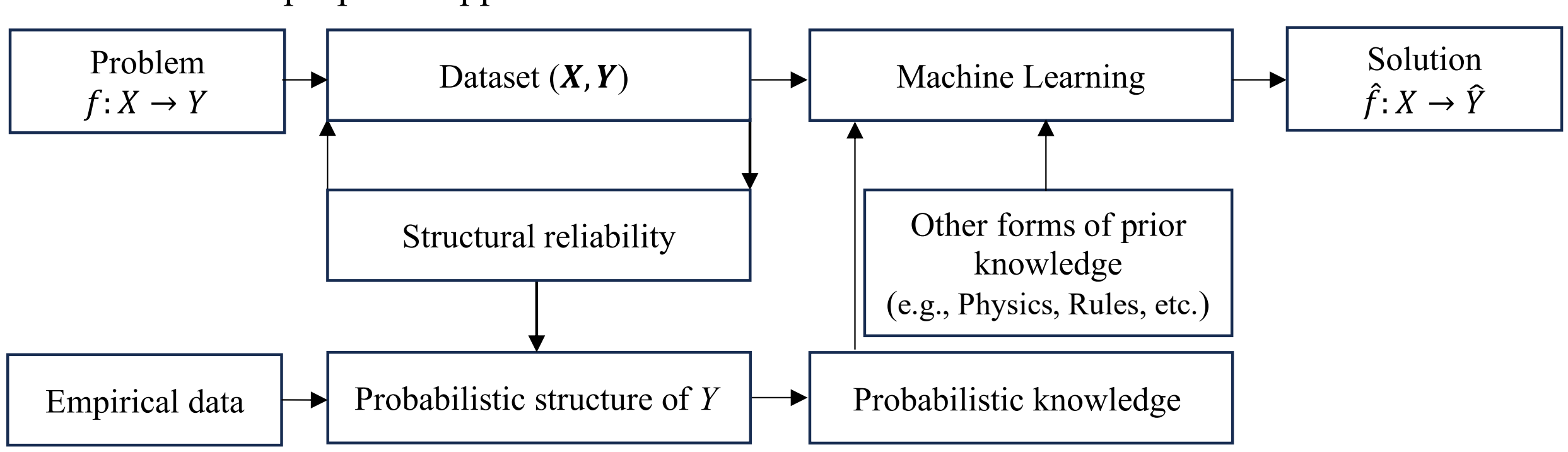

Fig. 6. The flow of information in the probabilities-informed machine learning

### 3.2. Pseudo-Algorithm and Architecture of Model

The required steps of the proposed approach are as follows:

1. Data Preparation and Distribution Estimation
   - Input: $\mathbf{X}$ (features), $Y$ (target).
   - If empirical data is available:
   - Derive $P_X(\boldsymbol{x})$ and $F_Y(y)$ using empirical observations.
   - If no empirical data is available:
   - Assume a probability distribution $P_X(\boldsymbol{x})$ for input $\mathbf{X}$ and use a reliability approach to estimate the CDF of $Y$ and augment the dataset by generating corresponding samples for $\mathbf{X}$.
2. Model Initialization
   - Define a machine learning model $f_\theta(X)$ with trainable parameters $\theta$.
   - Generate samples from $P_X(\boldsymbol{x})$, predict the output $f_\theta(X) = \hat{Y}$, and derive the CDF $\hat{F}_Y$.
   - Set the loss function to include:
     - $\mathcal{L}_{\text{Data}}(\theta)$ for point-wise accuracy (e.g., mean squared error).
     - $\mathcal{L}_{\text{Prob}}(\theta)$ for probabilistic consistency (e.g., $D_B\left(F_Y, \hat{F}_Y\right)$).
3. Loss Function Definition
   - Formulate the total loss: $\mathcal{L}(\theta) = \alpha \cdot \mathcal{L}_{\text{Data}}(\theta) + \beta \cdot \mathcal{L}_{\text{Prob}}(\theta)$.
     and any additional constraints derived from prior knowledge, if available.
4. Model Training
   - Optimize $\theta$ by minimizing the total loss $\mathcal{L}(\theta)$:
     $\theta^*= \arg\min_{\theta} \mathcal{L}(\theta)$
     Use an optimization method to update $\theta$.
5. Model Evaluation and Validation
   - Validate the model using test data:
     - Check point-wise accuracy with metrics like RMSE or MSE.
     - Verify probabilistic alignment by comparing $F_Y(y)$ and $\hat{F}_Y(y;\theta)$ using metrics like Bhattacharyya distance, KL divergence, Wasserstein distance, etc.

In the case of successful optimization, the trained model $f_\theta(X)$ predicts $Y$ with both point-wise accuracy and alignment to the true distribution $F_Y(y)$.

In this study, SVR and ANN are utilized to demonstrate how the proposed approach can enhance the prediction capabilities of learning machines.

Based on the proposed implementations, Fig. 7 illustrates the architecture of the neural network method, incorporating the probabilistic structure of Y as prior knowledge, where the weights $W$ and bias terms $B$ in the network are considered as hyperparameters. It is worth mentioning that the number of layers, neurons, and other architectural details can also be treated as parameters for optimization.

Table 1 presents some important hyperparameters for SVR as a learning machine. These parameters control the complexity, regularization, and kernel behavior of the model, which directly influence the model's predictive accuracy. Like the neural network-based approach, the learning process requires running the SVR model in each optimization process to derive the hyperparameters $\boldsymbol{\theta} =\{\mathcal{K}, \mathcal{B}, \epsilon, \mathcal{F}\}$ that minimize the cost in Eq. (4).

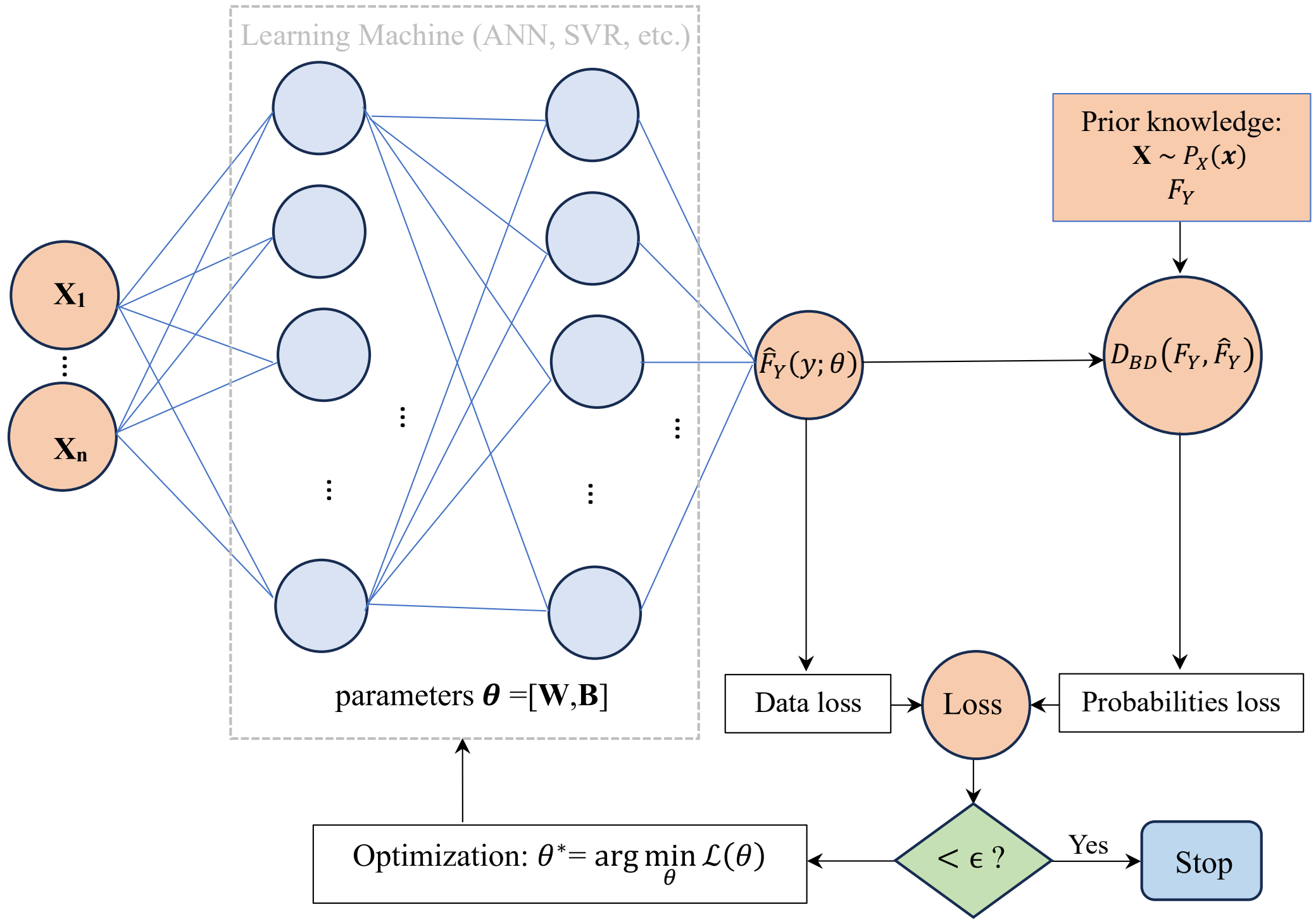

Fig. 7. Architecture of the Probabilities-Informed Neural Network. Other learning models (e.g., SVR with the hyperparameters listed in Table 1) can be embedded into the proposed framework as alternatives to Neural Networks

Table 1. Potential hyperparameters for SVR optimization using probabilities knowledge

| **Hyperparameter** | **Sign** | **Description** |
|---|---|---|
| KernelScale | $\mathcal{K}$ | Determines the kernel's feature mapping scale. |
| BoxConstraint | $\mathcal{B}$ | Regularization controlling the trade-off between training accuracy and model smoothness. |
| Epsilon | $\epsilon$ | Margin of tolerance for prediction errors. |
| KernelFunction | $\mathcal{F}$ | Kernel type for feature mapping. |

## 4. Verification and Performance Evaluation

To validate the proposed ML approach, we test it against several benchmark problems, encompassing both synthetic and real-world datasets. The Monte Carlo simulation was used to derive the CDFs while ANN and SVR were employed as the ML model $f_\theta(X)$. The optimization is performed by tuning the key hyperparameters $\theta$ using Adam and Bayesian optimization for ANN and SVR, respectively.

The capabilities of Probabilities-Informed Neural Networks (PRINN) are investigated in comparison to traditional ANNs for a Structural Health Monitoring (SHM) problem. Additionally, SVR optimized using the proposed approach is validated on benchmark image-denoising and classification problems.

### 4.1. Verification of PRINN

The proposed PRINN framework (see Fig. 7) is applied to a synthetic structural health monitoring (SHM) dataset. The goal is to predict damage levels under varying structural and environmental conditions. Datasets with varying sample sizes were synthetically generated. The inputs in these datasets included dynamic load amplitude ($F_D$), vibration frequency ($f_v$), material degradation factor ($M_D$), and crack density ($C_d$) represented measurable structural parameters, while the outputs corresponded to normalized damage levels.

For verification purposes, the damage level ($D$), can be synthesized as a function of the input variables in the normalized range of [-1, 1] with added Gaussian noise $\upsilon \sim \mathcal{N}(0, 0.1^2)$ to represent the real-world measurement variability:

$$D = Sin(\pi F_d) \cdot \log(|f_v| + 1) + 0.8 \cdot e^{-|M_d - 0.5|} \cdot cos(2\pi C_d) - 0.5 \cdot \frac{F_d \cdot f_v}{1 + {M_d}^2 + |C_d|} + 0.2 \cdot \sin(10 \cdot F_d) \cdot \sin(5 \cdot f_v) + \upsilon. \quad (13)$$

To model the nonlinear relationship between the input features and the damage levels, a feedforward neural network with two hidden layers, each with 120 neurons, followed by a hyperbolic tangent (tanh) activation function to introduce nonlinearity. The network was trained using a total of 1500 iterations with a learning rate of 0.01 for both the baseline model and the proposed approach. The loss function of Eq. (4) is defined as:

$$\mathcal{L}(\theta) = \alpha \cdot \mathcal{L}_{\text{Data}}(\theta) + (1 - \alpha) \cdot D_{JS}(F_Y, \hat{F}_Y), \quad (14)$$

where $\mathcal{L}_{\text{Data}}(\theta)$ represents the data-driven loss, and $D_{JS}(F_Y, \hat{F}_Y)$ is the Jensen-Shannon Divergence between the predicted and actual cumulative distributions (see Section 2.2). The parameter Alpha ($\alpha$) is considered dynamic with respect to iterations as shown in Fig 8.

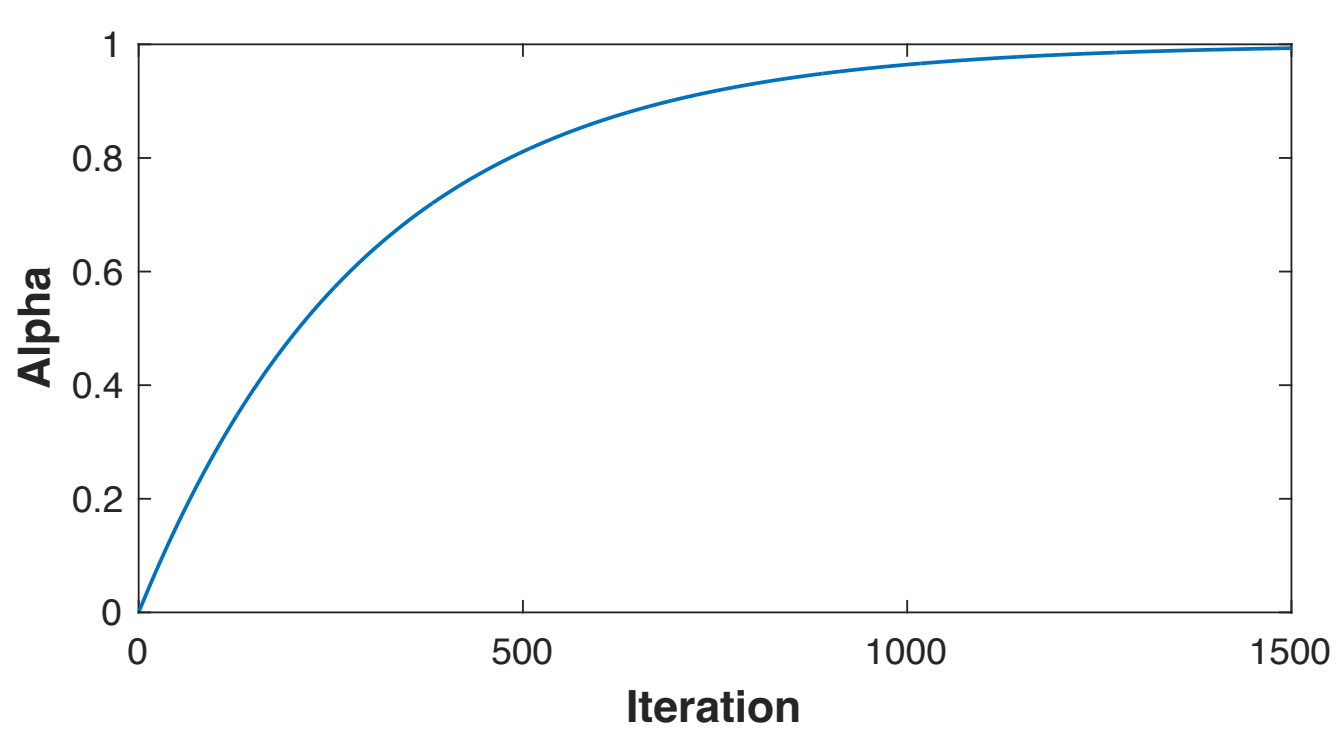

Fig. 8. Dynamic parameter Alpha ($\alpha$): Adapting the weighting factor to balance probability and data loss during training

Assuming that the probability distribution of input variables and CDF information on damage level from empirical observation is available, we trained ANN with different input samples *n* as *n*=5 and *n*=100.

In the case of *n*=5, due to the very small input size, the baseline model achieves near-zero training error but struggles with poor generalization, as shown in Fig. 9. This failure to capture the behavior of the original model highlights significant overfitting. In contrast, the probabilities-informed method achieves near-perfect performance by emphasizing alignment between the predicted and actual distributions (see Fig. 10). This alignment is crucial for accurately capturing uncertainty in real-world applications. Notably, training a model with only 5 samples while achieving a correlation coefficient of 92% underscores the importance of incorporating probabilistic information into the training process of PRINN.

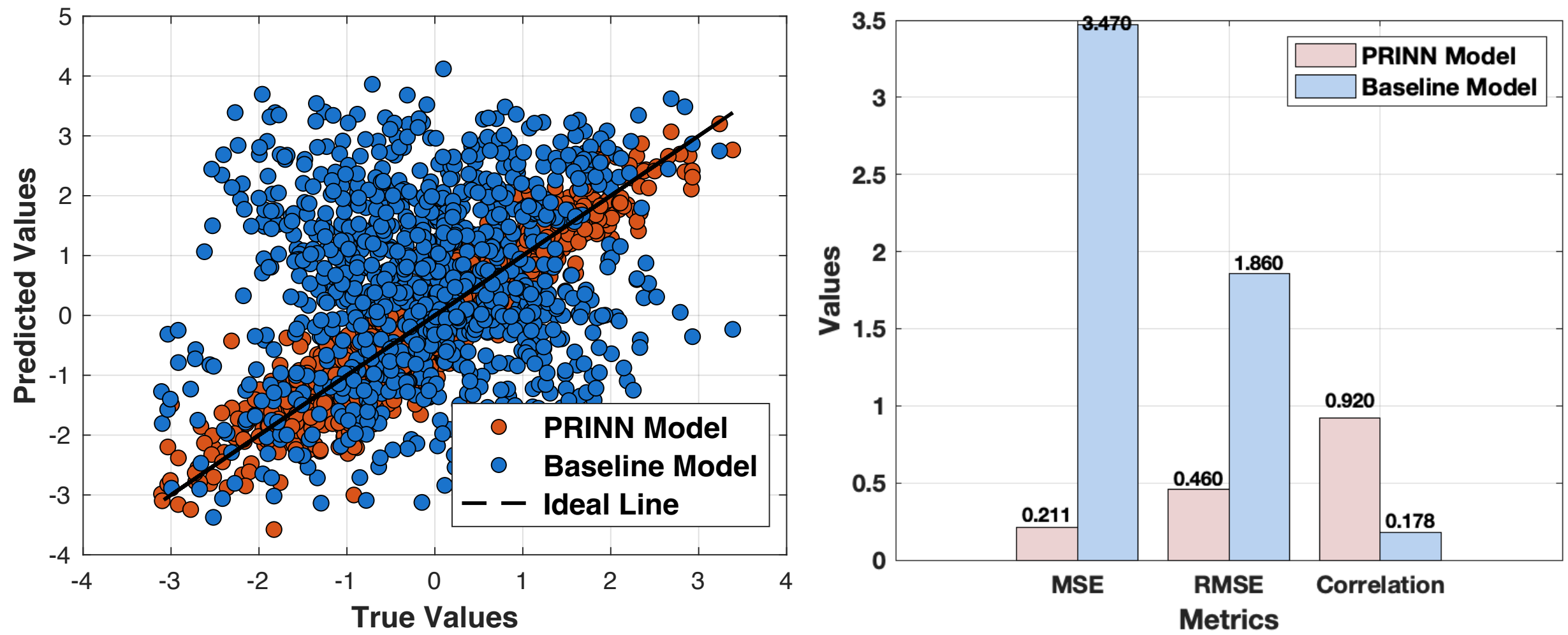

Fig. 9. Comparison of performance metrics between the proposed PRINN approach and baseline models in the structural health monitoring example with $n$=5

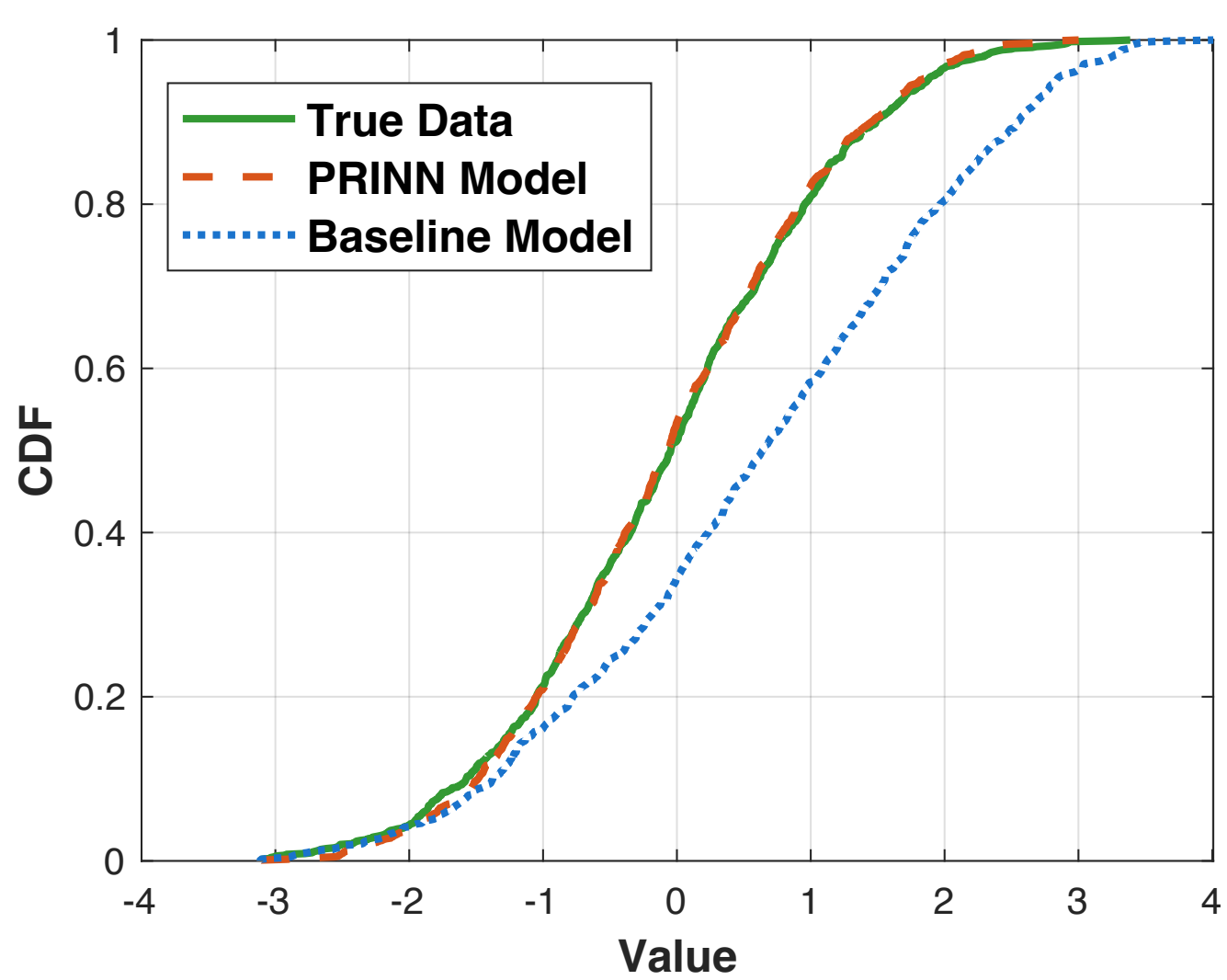

Fig. 10. Comparison of output CDFs between the proposed PRINN approach and baseline models in the structural health monitoring example with $n$=5

Fig. 11 compares the performance metrics of the proposed PRINN approach and baseline models for $n$=100. The results indicate that increasing the input sample size improves the accuracy of both models. For the baseline model, the correlation coefficient increases to 58%, reflecting some improvement in capturing the probabilistic information of Y (See Fig 12). Nonetheless, given the low correlation coefficient, the model's accuracy may not be adequate for real-world applications.

In contrast, the PRINN approach achieves a correlation coefficient of 96.6%, demonstrating the high accuracy of the model when a balanced number of input samples and probabilistic information of the output Y are utilized.
These findings emphasize the utility of the probabilities-informed method in scenarios where capturing the probabilistic characteristics of predictions is critical. By integrating probabilistic metrics during optimization, the proposed approach effectively balances pointwise error minimization and distributional fidelity, providing a significant advantage over conventional methods in applications requiring robust representation of uncertainty.

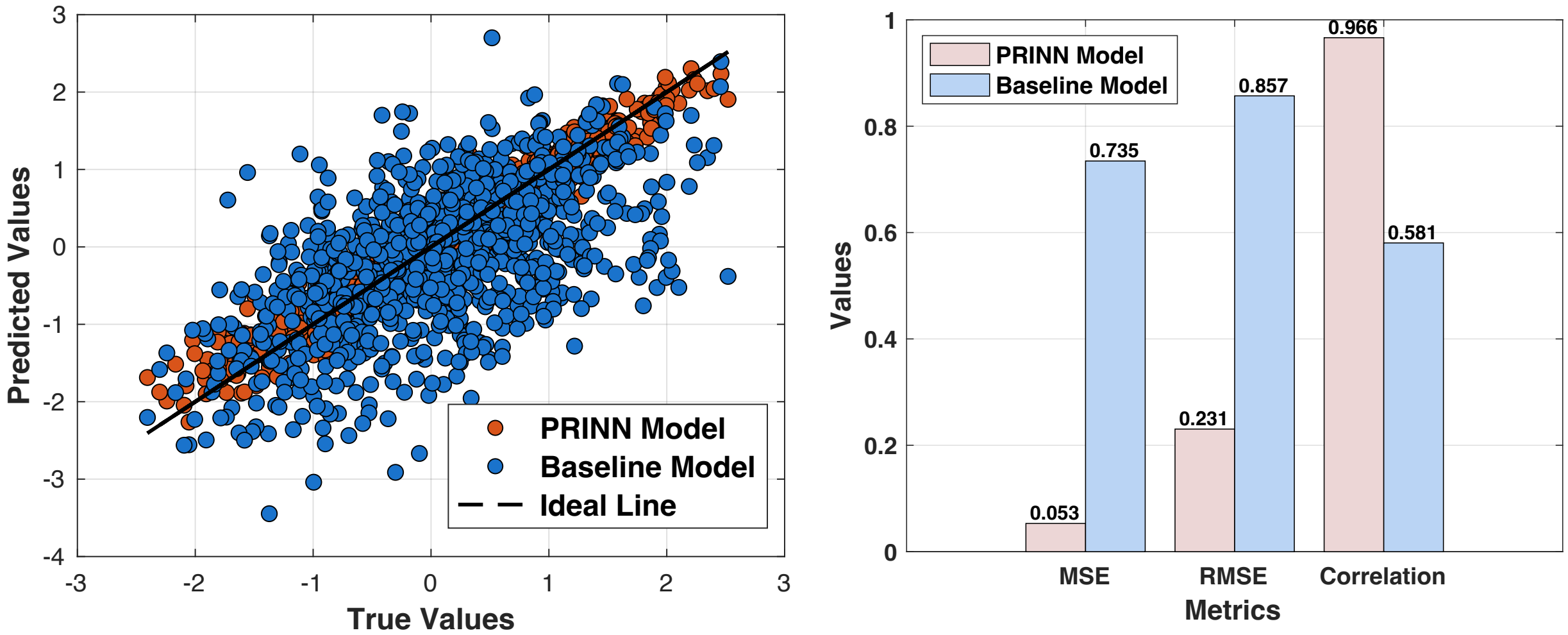

Fig. 11. Comparison of performance metrics between the proposed PRINN approach and baseline models in the structural health monitoring example with $n$=100

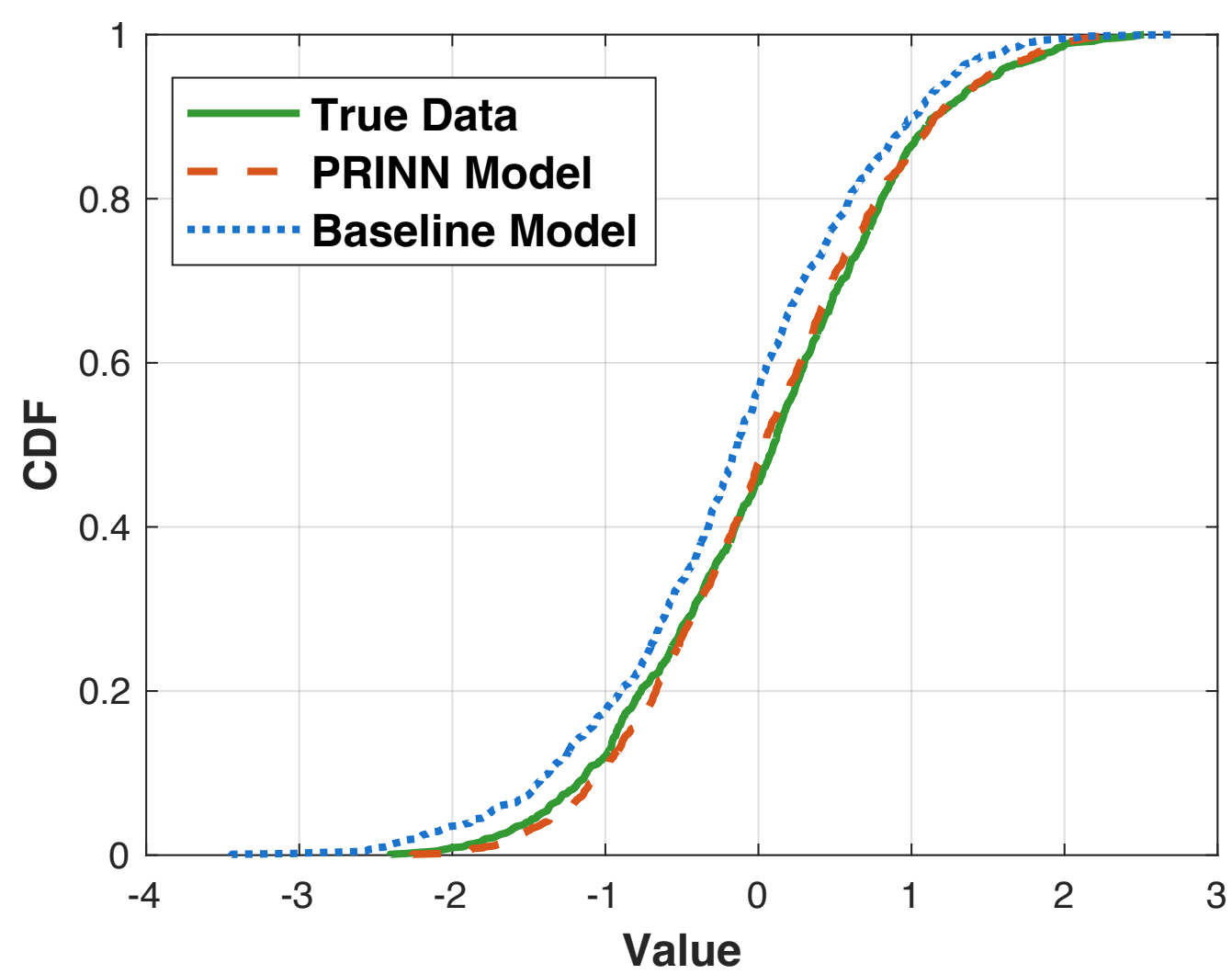

Fig. 12. Comparison of output CDFs between the proposed PRINN approach and baseline models in the structural health monitoring example with $n$=100

### 4.2. Verification of Probabilities-Informed Support Vector Machine

This section explores the scenario where SVR is used as the learning machine. For verification purposes, we compare the performance of three SVR models with the same training points:

1. Baseline model (SVR with default parameters): This model is trained using the standard hyperparameter settings provided by Matlab's SVR implementation (version 2024a). It serves as a reference to assess the improvements achieved by the proposed approach.
2. Proposed probabilities-informed approach: This model leverages a custom loss function defined as Eq. (4).
3. RMSE-optimized approach: A model trained solely to minimize the data-driven loss term
$$\mathcal{L}(\theta) = \mathcal{L}_{\text{Data}}(\theta) \quad (15)$$
This scenario assesses the impact of incorporating probabilistic information on model performance.

#### 4.2.1. Image Denoising Example

In this example, we evaluate the effectiveness of the probabilities-informed approach in image denoising. The denoising task involves reconstructing the original (noise-free) image from its noisy counterpart. To evaluate the proposed method, we use the "Cameraman" image, a well-known image available in Matlab's library. Gaussian noise with a mean of zero and a standard deviation of 0.1 ($\nu \sim N(0,0.1^2)$) is added to the image to simulate real-world noise. The original and noisy images are shown in Fig. 13.

The task involves using the proposed probabilities-informed trained model to remove noise from images and comparing the results with two alternatives: a baseline SVR model trained using default Matlab settings, and a model optimized solely based on a data-driven loss function.

Instead of estimating the CDF of the output $Y$ (which may not be directly available for the original image) during the training process, we estimate the CDF of the training samples. The objective is to identify a model that not only achieves accurate denoising but also captures the probabilistic structure inherent in the training samples. This allows us to demonstrate how the inclusion of probabilistic optimization in the proposed approach improves the image quality compared to traditional methods. In this denoising process, the proposed probabilities-informed model leverages a custom loss function defined as: $\mathcal{L}(\theta) = 0.3 \cdot \mathcal{L}_{\text{Data}}(\theta) + 0.7 \cdot D_B\left(F_Y, \hat{F}_Y\right)$.

The effectiveness of the denoising is assessed using two key metrics:

- PSNR (Peak Signal-to-Noise Ratio): A metric that compares the quality of the denoised image, where higher values indicate better image quality.
- SSIM (Structural Similarity Index): This measures how similar the denoised image is to the original image, focusing on structural content.

We have used the Bayesian optimization approach to obtain hyperparameters of SVR. The denoising results and corresponding metrics are presented and plotted in Figs. 14 and 15.

The PSNR for the probabilities-informed SVR model is higher compared to the baseline SVR and RMSE-optimized model, indicating that the denoised image has fewer distortions and artifacts. Similarly, the SSIM for the probabilities-informed SVR model is also higher than the other two models, demonstrating that the denoised image preserves structural details better. This metric is crucial because it focuses on the perceptual quality of the image, and our approach provides better structural fidelity than the other models.

The results reveal the potential of the proposed approach for image-denoising problems, demonstrating that probabilistic metrics can be seamlessly integrated into the objective function.

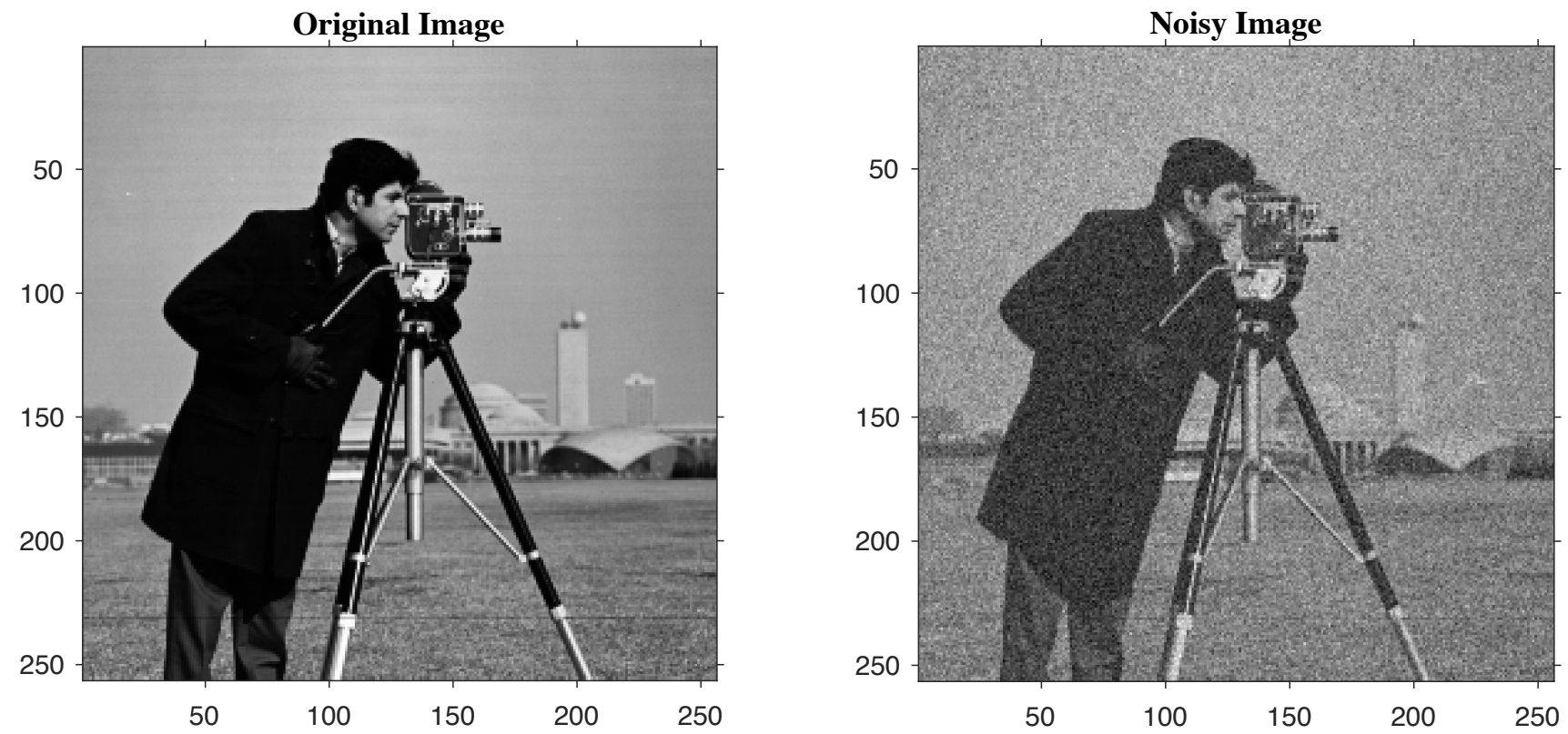

Fig. 13. The original and noised image of example 2

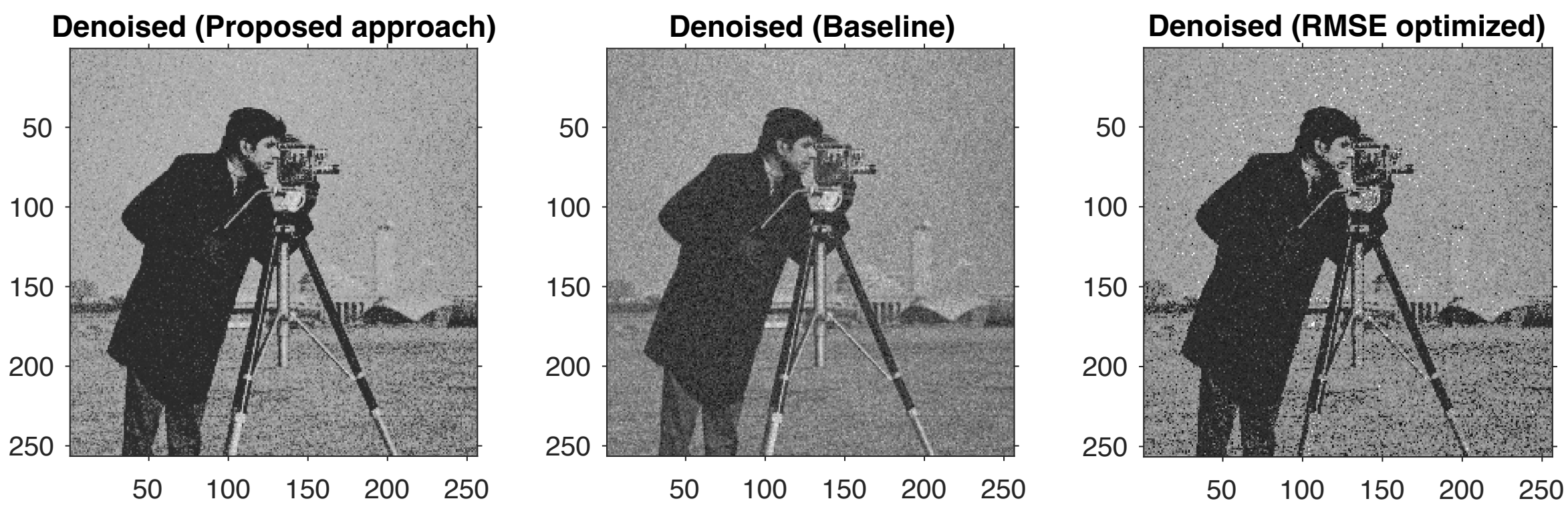

Fig. 14. The denoised images of three SVR models

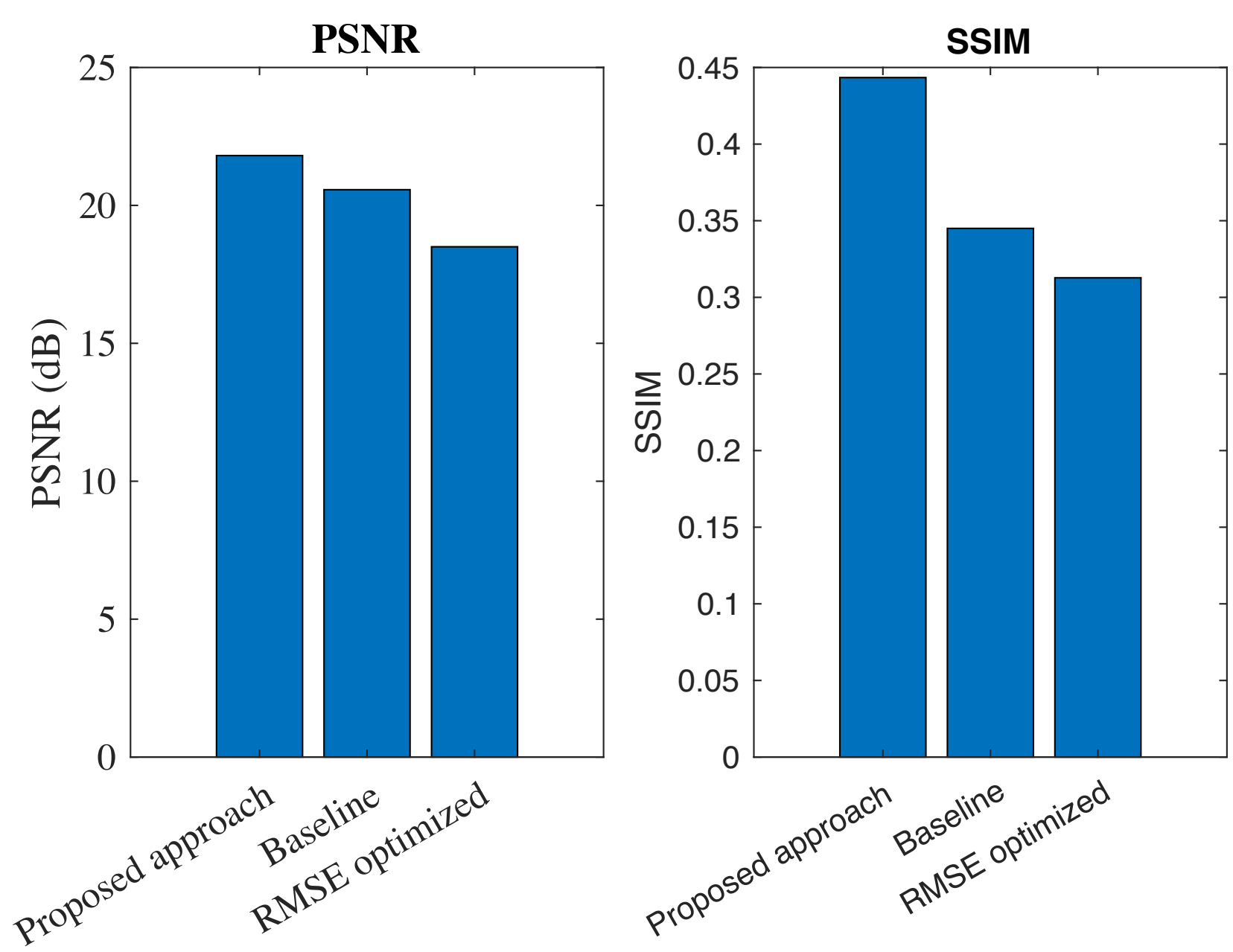

Fig. 15. The denoising metrics of three SVR models

#### 4.2.2. Moderate/High Dimensional Classification Example

As the last example, we investigate the classification performance of the proposed approach on the "Ionosphere" dataset, which is a well-known dataset in ML. The corresponding dataset contains 351 instances, each described by 34 numeric features, and the target variable is binary. The objective is to train classification models to predict whether the radar return is from the ionosphere (good) or not (bad). The dataset has an approximately balanced distribution of the two classes, which makes it suitable for classification tasks.

The problem is firstly solved by the three Support Vector Classification (SVC) models while 80% of the data is used for testing, and the remaining 20% is used for training (poor $\mathbf{X}$ and $Y$ dataset). The obtained confusion matrix and classification error of trained models are reported in Figs 16-17, and more accurate discussions of results, and performance metrics for three considered models based on the confusion matrix are summarized in Table 2.

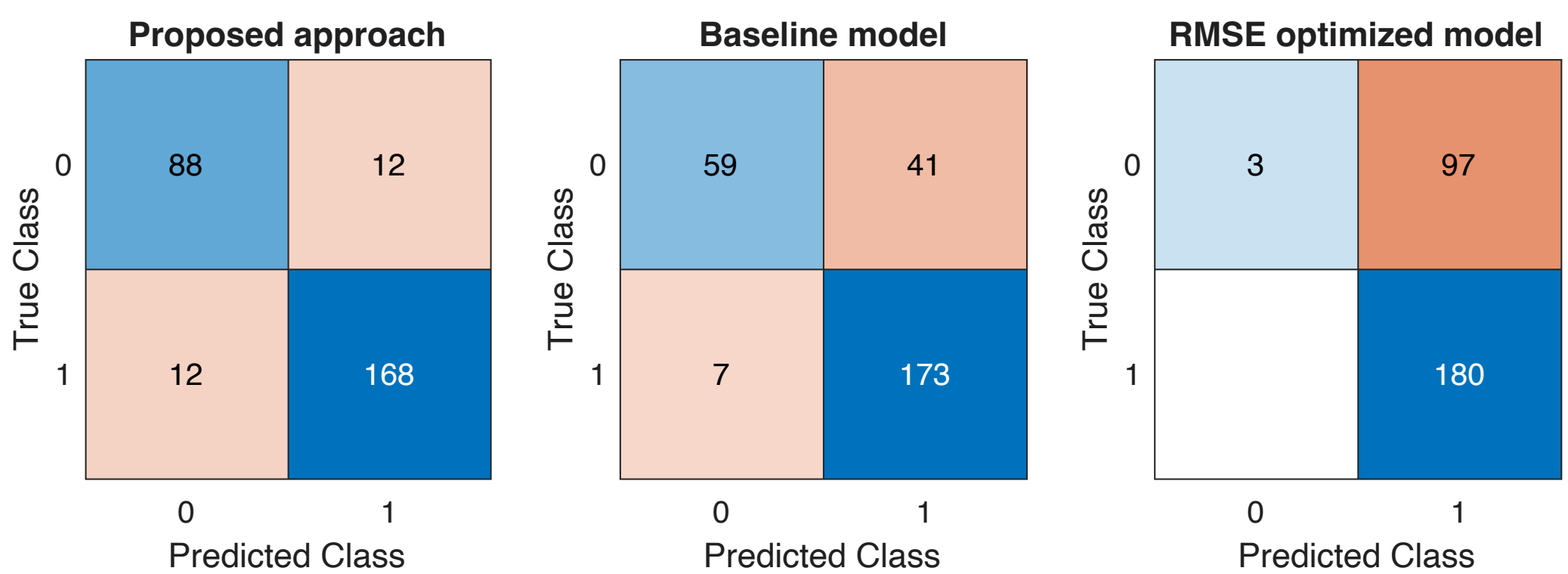

Fig. 16. Confusion matrix of the three SVC models for the Ionosphere dataset

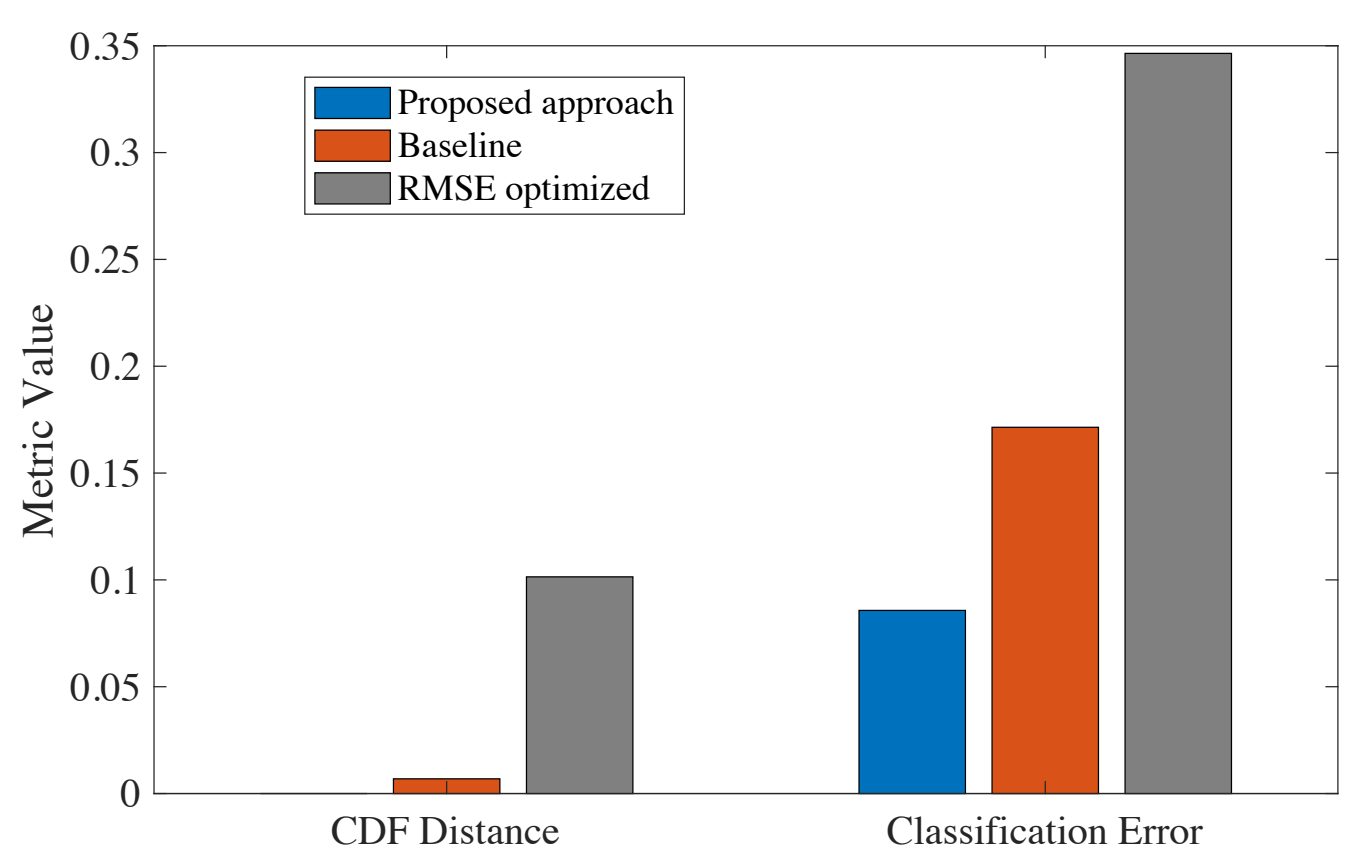

Fig. 17. Classification error and CDF distance of the three SVC models for the Ionosphere dataset

Table 2. Performance metrics of three SVC models

| Model | Accuracy | Precision | Recall | F1-Score |
|---|---|---|---|---|
| Proposed approach | 0.91 | 0.90 | 0.91 | 0.91 |
| Baseline | 0.83 | 0.85 | 0.78 | 0.79 |
| RMSE optimized | 0.65 | 0.82 | 0.51 | 0.42 |

In the case of the Accuracy metric, the baseline SVC model achieves an accuracy of 0.8286, while the RMSE optimized model achieves a much lower accuracy of 0.6536, indicating that optimizing for RMSE alone does not lead to good classification results and fails to align to maximize classification performance. In comparison, the proposed probabilities-informed model demonstrates a notable improvement with an accuracy of 0.9143, marking an increase of nearly 8% over the default model. This suggests that the optimization method significantly enhances the model's ability to classify instances correctly.
In the case of the Precision metric, the Baseline model achieves a precision of 0.8512, indicating that 85.12% of the instances predicted as positive are correctly classified as good. However, the RMSE-optimized model performs less effectively with a precision of 0.8249, suggesting that (as in the former case) focusing on RMSE optimization is not ideal for improving precision in classification tasks. The proposed probabilities-informed model improves precision to 0.9067, meaning that it better identifies positive instances as true positives.
In this example, the Recall metric for the Baseline model is 0.7756, meaning it correctly identifies 77.56% of the actual positive instances (good). The proposed probabilities-informed model shows a substantial improvement in recall, reaching 0.9067, which means it correctly identifies 90.67% of all actual positive instances. In contrast, the RMSE-optimized model significantly underperforms with a recall of 0.5150, missing a large proportion of the positive instances.
The F1-score for the Baseline model is 0.7945, reflecting a reasonable balance between precision and recall. The proposed probabilities-informed model achieves a higher F1-score of 0.9067, indicating an improved balance between precision and recall. However, the RMSE-optimized model performs poorly with an F1-score of 0.4230, further confirming that optimizing for RMSE is not suitable for classification tasks where balancing precision and recall is essential.
As the second try, 50% of the data is used for testing, and the remaining 50% is used for training. The results of the confusion matrix are presented in Fig 15. Results show that the proposed model outperforms the other models across all evaluation metrics, including accuracy, precision, recall, and F1-score. These results demonstrate the effectiveness of the proposed optimization approach in improving classification performance on the Ionosphere dataset.

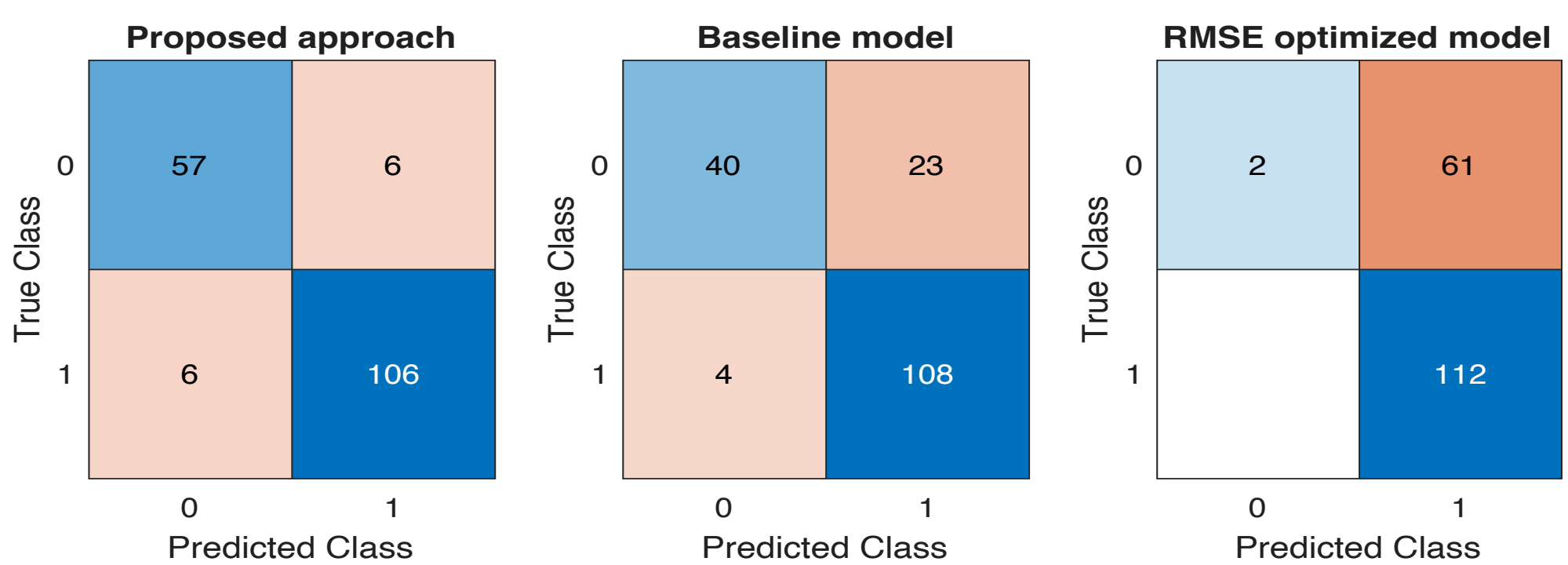

Fig. 15. Confusion matrix of three SVC models for the Ionosphere dataset with 50% training input

## 5. Conclusion

In this study, we demonstrated that integrating the probabilistic characteristics of the output function $Y$ (e.g., the probability density function) into the training process of machine learning (ML) models significantly enhances model performance compared to traditional methods. This approach, inspired by physics-informed ML, utilizes available probabilistic insights from real-world data or estimates obtained through structural reliability methods during experimental design. In the case of using reliability methods, by assuming that the input variable X follows a specific cumulative distribution function (CDF), the corresponding CDF of $Y$ can be derived (recognizing that estimating very small probabilities is unnecessary in this context, especially when precise prediction of extreme values is not essential), various numerical and approximate methods from structural reliability theory can be utilized to estimate $F_Y(y)$). Consequently, predictive models should ensure alignment between the predicted and original output CDFs for X with the specified distribution.

To achieve this, the proposed framework optimizes ML models such (e.g., Artifical Neural Network, Support Vector Machine, and etc.) and by minimizing divergence measures like Bhattacharyya distance between predicted and actual output distributions. Incorporating such probabilistic knowledge mitigates overfitting and underfitting by aligning model predictions with the statistical properties of the target variable. Real-world examples highlight the method's effectiveness, demonstrating that embedding the probabilistic structure of outputs fosters improved generalization and robustness. These findings underscore the practical value of the proposed approach, establishing it as a robust alternative for enhancing ML model reliability in diverse applications.